\documentclass{article}

\usepackage{PRIMEarxiv}

\usepackage[utf8]{inputenc} 
\usepackage[T1]{fontenc}    
\usepackage{hyperref}       
\usepackage{url}            
\usepackage{booktabs}       
\usepackage{amsfonts}       
\usepackage{nicefrac}       
\usepackage{microtype}      
\usepackage{lipsum}
\usepackage{graphicx}
\graphicspath{{media/}}     

\title{ %
    {\rmfamily{\scshape{TailorMe}}}: Self-Supervised Learning of an \\%
    Anatomically Constrained Volumetric Human Shape Model %
}

\author{
  Stephan Wenninger, Fabian Kemper\\
  TU Dortmund University \\
  Dortmund \\
  \texttt{\{stephan.wenninger, fabian.kemper\}@tu-dortmund.de} \\
  \And
  Ulrich Schwanecke \\
  Hochschule RheinMain \\
  Wiesbaden \\
  \texttt{ulrich.schwanecke@hs-rm.de} \\
  \And
  Mario Botsch \\
  TU Dortmund University \\
  Dortmund \\
  \texttt{mario.botsch@tu-dortmund.de} \\
}

\graphicspath{{./figures/}}

\usepackage[separate-uncertainty = true,multi-part-units=single]{siunitx}
\usepackage{amsmath}

\renewcommand{\vec}[1]{\mathbf{#1}}
\newcommand{\mat}[1]{\mathbf{#1}}
\newcommand{\set}[1]{\mathcal{#1}}

\newcommand{\N}{\mbox{\rm \hbox{I\kern-.15em\hbox{N}}}}
\newcommand{\R}{\mathbb{R}}
\newcommand{\laplace}{\mathrm{\Delta}}

\newcommand{\of}[1]{\!\left( #1 \right)}
\newcommand{\abs}[1]{\left| #1 \right|}
\newcommand{\norm}[1]{\left\Vert {#1} \right\Vert}
\renewcommand{\matrix}[1]{\begin{pmatrix} #1 \end{pmatrix}}

\newcommand{\T}{^\mathsf{T}}

\newcommand{\Sec}[1]{Section~\ref{sec:#1}}
\newcommand{\Fig}[1]{Figure~\ref{fig:#1}}

\newcommand{\Eq}[1]{Equation~\eqref{eq:#1}}

\newcommand{\template}[1]{\tilde{#1}}
\newcommand{\skin}{\mathcal{S}}
\newcommand{\bone}{\mathcal{B}}
\newcommand{\pskin}{\left(\skin\right)}
\newcommand{\pbone}{\left(\bone\right)}
\newcommand{\tetmesh}[1]{\mathbb{#1}}
\newcommand{\highresskel}{\mathcal{SK}}
\newcommand{\collisions}{\mathcal{C}}

\newcommand{\excluded}{\mathcal{E}}

\begin{document}

\maketitle

\begin{figure}[!h]
    \centering
    \includegraphics[width=1.0\linewidth]{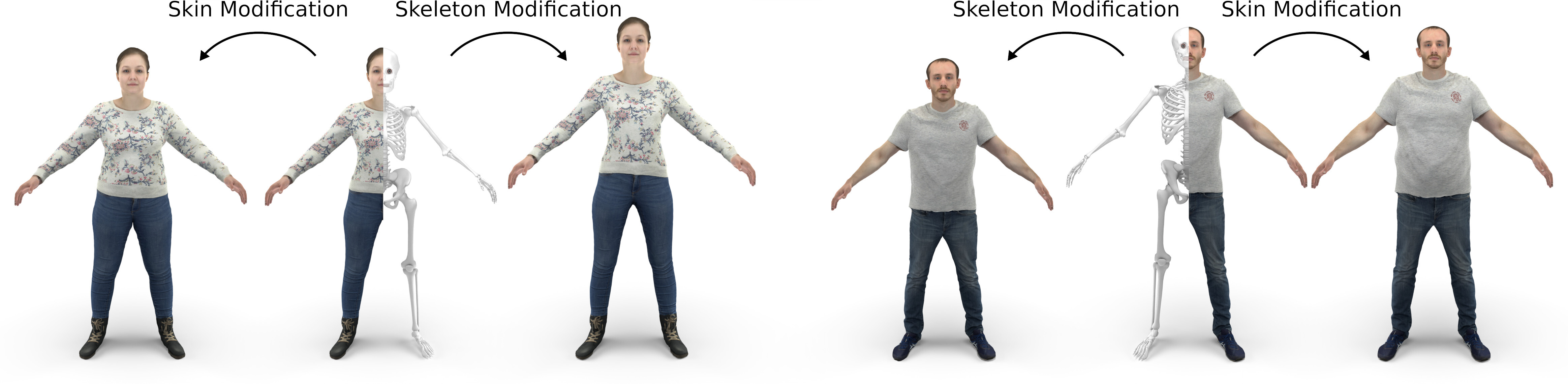}
    \caption{{\scshape TailorMe}: Our anatomically constrained volumetric human shape model allows to infer skeleton shape from a given surface scan. Due to injecting anthropometric measurements into the latent code of our model, we can then locally manipulate both the skeleton shape and the soft tissue distribution of a given person.}
    \label{fig:teaser}
\end{figure}

\begin{abstract}
Human shape spaces have been extensively studied, as they are a core element of human shape and pose inference tasks. Classic methods for creating a human shape model register a surface template mesh to a database of 3D scans and use dimensionality reduction techniques, such as Principal Component Analysis, to learn a compact representation. While these shape models enable global shape modifications by correlating anthropometric measurements with the learned subspace, they only provide limited \emph{localized} shape control. We instead register a volumetric anatomical template, consisting of skeleton bones and soft tissue, to the surface scans of the CAESAR database. We further enlarge our training data to the full Cartesian product of all skeletons and all soft tissues using physically plausible volumetric deformation transfer. This data is then used to learn an anatomically constrained volumetric human shape model in a self-supervised fashion. The resulting {\scshape TailorMe} model enables shape sampling, localized shape manipulation, and fast inference from given surface scans.
\end{abstract}

\keywords{cs.GR \and Mesh models \and Volumetric models \and Shape analysis \and Learning latent representations}


\section{Introduction}

Human shape modeling has been extensively studied due to its application in various fields, such as shape and pose estimation from multi-view stereo or monocular RGB(-D) input.
Starting from simple linear PCA models \cite{Anguelov2015Scape, Loper2015SMPL} to recent advances in machine learning models \cite{ranjan2018coma, bouritsas2019neural3dmm}, these models are used as the foundation for many downstream tasks such as body composition estimation \cite{wong2021bodycomposition}, the creation of virtual humans \cite{achenbach17, Alldieck2019Octopus, Wenninger2020}, or generating synthetic training data for image recognition tasks \cite{Wood2021_FakeIt}.
Most of the approaches train on commercially available 3D scan databases such as CAESAR \cite{robinette2002caesar} or 3D Scanstore \cite{3dscanstore}.
These 3D scans naturally provide only what is easily observable from the outside: the silhouette of the scanned subject.
However, by setting the focus on modelling the \emph{skin layer} of humans, models that want to learn how to accurately \emph{modify} a given virtual human, suffer from missing anatomical information.

Modifying realistic virtual humans has gained attention due to its promising applicability in VR therapy \cite{piryankova2014, moelbert2018assesing, wolf2022distance, wolf2022plausibility}, which can serve as a complementary intervention technique to classic forms of therapy.
Such VR systems can immersively expose patients with anorexia or obesity to generic or personalized virtual humans at different levels of weight or Body Mass Index (BMI), allowing patients to reflect on and researchers to gain insight into possibly occurring body image disturbances.  
However, current models for body weight/BMI modification are typically learnt on surface models only \cite{allen06learning, hasler2009, piryankova2014, dollinger2022resizeme}, leading to shape modification models providing only limited localized control. 
Participants have stated the request for changing the composition of specific body parts in addition to a global BMI/weight modification \cite{dollinger2022resizeme}.

In this paper, we present our {\scshape TailorMe} model, which is able to achieve such localized shape manipulation.
We leverage recent advances in inferring anatomical structures from surface scans \cite{dicko2013anatomy, kadlecek16, komaritzan2021inside, Keller2022OSSO} to register a volumetric anatomical template model to the CAESAR database, resulting in pairs of skeleton and skin meshes. 
In order for our model to successfully learn separate parameter sets for skeleton shape and soft tissue distribution, we calculate the full Cartesian product by means of volumetric deformation transfer \cite{botsch2006deftrans}. 
This allows us to transfer the soft tissue distribution of subject $i$ onto the skeleton of subject $j$ in a physically plausible way.

The resulting data set is then used to train a neural network that learns the common underlying parameters from a person's bone structure and soft tissue distribution. Examples that share either a common skeleton or a common soft tissue distribution can be sampled from the Cartesian data set. These commonalities are learned and encoded with an autoencoder using the SpiralNet++\cite{gong2019spiralnetplus} approach.
To separate the skeleton and soft tissue distribution a self-supervised learning approach inspired by Barlow Twins \cite{zbontar2021barlow} is used, which reduces redundancy in the underlying distributions. 
To allow local modification of body regions, measurements are taken on the example Cartesian data set and are additionally injected into the latent code to reduce the correlation of the remaining parameters with these known measurements. 

In summary, this paper presents an approach for learning an anatomically constrained volumetric human shape model, whose latent code can be sampled to generate various human skeleton shapes with different soft tissue distribution characteristics. 
The measurement injection into the latent code of our model allows localized shape manipulation: the anatomical structure can be quickly inferred from a 3D scan of a human and then locally modified by the user.
This allows to simulate weight gain/loss in different regions of the body. 
We will publicly release our {\scshape TailorMe} model to enable further research and development of applications for volumetric anatomical human shape models.

\section{Related Work}

\subsection{Human Shape Models}

Data-driven human shape models are ubiquitous and widely studied. Learnt from registering a template model to a database of 3D scans, most popular models are based on Principal Component Analysis (PCA) on vertex positions \cite{Loper2015SMPL, Osman2020STAR}.
Other approaches directly encode triangle deformations from the template to the registered models \cite{Anguelov2015Scape} or a decomposition of these triangle deformations \cite{Freifeld:ECCV:2012}. 
Since they are based on a database of 3D scans, these methods capture the variation of human body shape only on a surface level. 
In contrast, our model is trained on additional volumetric information by fitting an anatomically plausible skeleton model into the registered surface scans.

More sophisticated dimensionality reduction techniques have also been applied to human shape models: 
Ranjan et al.~\cite{ranjan2018coma} propose a convolutional mesh autoencoder and introduce a pooling and unpooling operation directly on the mesh surface structure. 
The Neural 3D Morphable Models (Neural3DMM) network \cite{bouritsas2019neural3dmm} adjusts the pooling operations and uses a spiral convolutional operator, which has been further refined by Gong et al.~\cite{gong2019spiralnetplus}. 
Our model uses a similar autoencoder design paired with the self-supervised learning technique Barlow Twins \cite{zbontar2021barlow}.

\subsection{Modifying Virtual Humans}

Learning a shape modification model based on anthropometric measurements has been explored in the field of Virtual Reality body image therapy \cite{dollinger2022resizeme, wolf2022distance, wolf2022plausibility, piryankova2014, moelbert2018assesing}.
The possibility to either passively present a generic virtual human in different weight or BMI variants or letting participants actively change their personalized virtual human can and has been used to gain insights into body image disorders for patients with anorexia or adiposity.

A common approach is to model shape modification by learning linear correlations between a set of anthropometric measurements (e.g., as present in the CAESAR database \cite{robinette2002caesar}) and the low-dimensional shape space \cite{allen06learning, hasler2009, piryankova2014, dollinger2022resizeme}.
The modified shape can then be computed by mapping the desired measurement changes into the subspace through learned regressors and then projecting the change in subspace coordinates back into vertex space. Commonly used anthropometric measurements, such as arm length and inseam, are highly correlated.
The cited methods cannot completely disentangle this correlation in the anthropometric measurements, leading to limited control over local shape manipulation. 
Our non-linear model learns to separate the correlations between these measurements, thereby enabling more localized shape manipulations.

\subsection{Anatomical Models}

Achenbach et al.~\cite{achenbach2018} trained a multi-linear model (MLM) to find a lower-dimensional model of skull and corresponding head shape, parameterized by skull shape and soft tissue distribution.
The MLM does not completely decouple the two parameter sets, so changing the skin parameters can still affect the skeleton.
Our non-linear model better decouples skeleton from skin shape, i.e., when changing skin parameters, the skeleton stays fixed.

Anatomy Transfer \cite{dicko2013anatomy} is a method for warping an anatomical template model into a target skin surface via a harmonic space warp while constraining bones to only deform via affine transformations. 
This can however lead to unnaturally scaled or sheared bones, as discussed in other work \cite{kadlecek16, komaritzan2021inside}.

Saito et al.~\cite{saito15} developed a physics-based simulation of muscle and fat growth on a tetrahedral template mesh including an enveloping muscle layer that separates the tetrahedral mesh representing the subcutaneous fat layer from the rest of the template.
\cite{kadlecek16} present a method for fitting such a physics-based simulation to a set of 3D scans in different poses, to get a personalized anatomical model.
Their approach yields visually plausible results but requires a complex numerical optimization strategy taking several minutes and can therefore not be used for interactive VR interventions.
Komaritzan et al.~\cite{komaritzan2021inside} follow the approach of Saito et al.~\cite{saito15} by using a multi-layered model to separate skeleton, muscle, and skin surface derived from an anatomical template model \cite{zygote}.
Their model is then fitted to a given skin layer in a multi-stage optimization scheme.
Embedding the high-resolution skeleton and muscle meshes from the anatomical template into the resulting layers is done by a triharmonic RBF warp.
However, they do not train a statistical model on the resulting shapes.
Additionally, their fitting approach is an order of magnitude slower compared to our method.

The recent work \emph{OSSO} \cite{Keller2022OSSO} combines the \emph{STAR} model \cite{Osman2020STAR} for human body shapes and a model of skeleton shapes based on the Stitched Puppet Model \cite{Zuffi:CVPR:2015}.
By fitting these two models to a set of DXA images, the authors learn to regress skeletal shape from skin shape in PCA space.
The skeletons inferred with the OSSO approach may however show self-intersections with the given skin mesh.
Our model learns non-linear correlations between skeleton and skin, provides a localized shape modification model, and produces intersection-free pairs of skeleton and skin meshes.

\section{Training Data}
\label{sec:training_data}

We start by deriving all the parts of our template model and registering it to surface scans of the CAESAR database, yielding pairs of skeleton and skin meshes (\Sec{template_and_caesar}). We enlarge this data set by computing the full Cartesian product of skeleton shape and soft tissue distribution via volumetric deformation transfer (\Sec{vol_deftrans}). 
The resulting data set then constitutes the training data for our {\scshape TailorMe} model.



\subsection{Skin and Skeleton Registration}
\label{sec:template_and_caesar}

\begin{figure}
    \centering
    \includegraphics[width=0.29\linewidth]{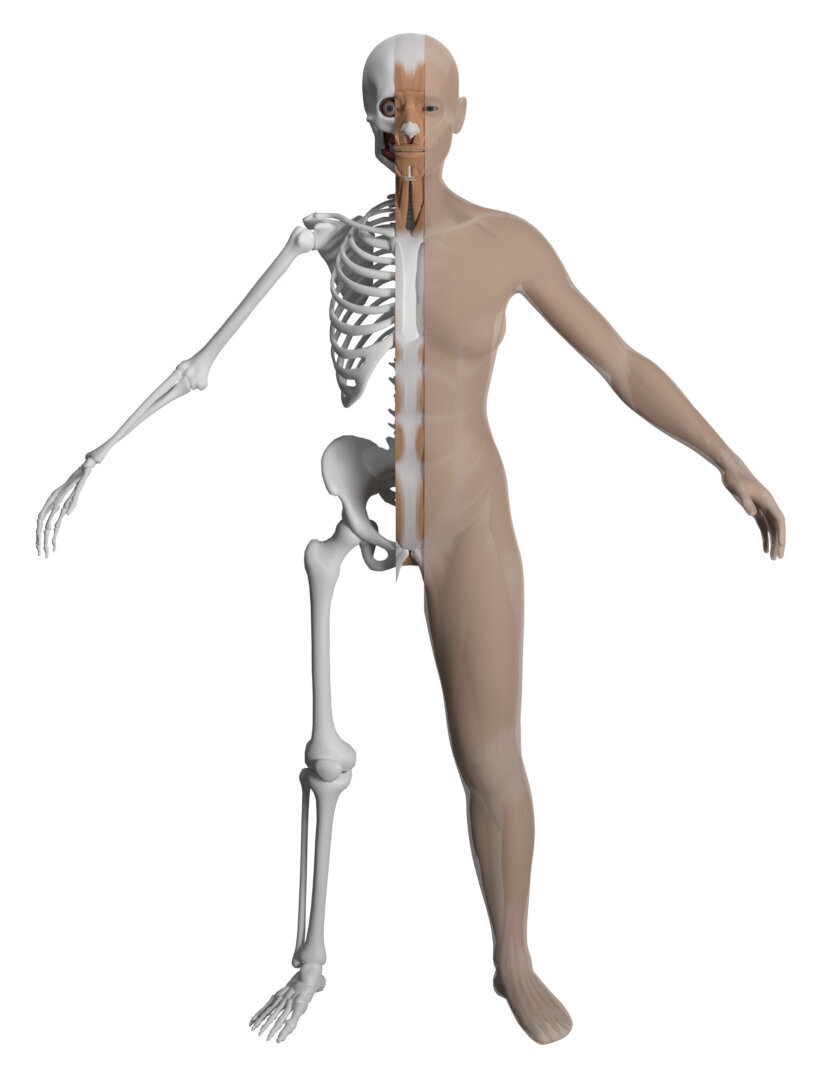}
    \includegraphics[width=0.29\linewidth]{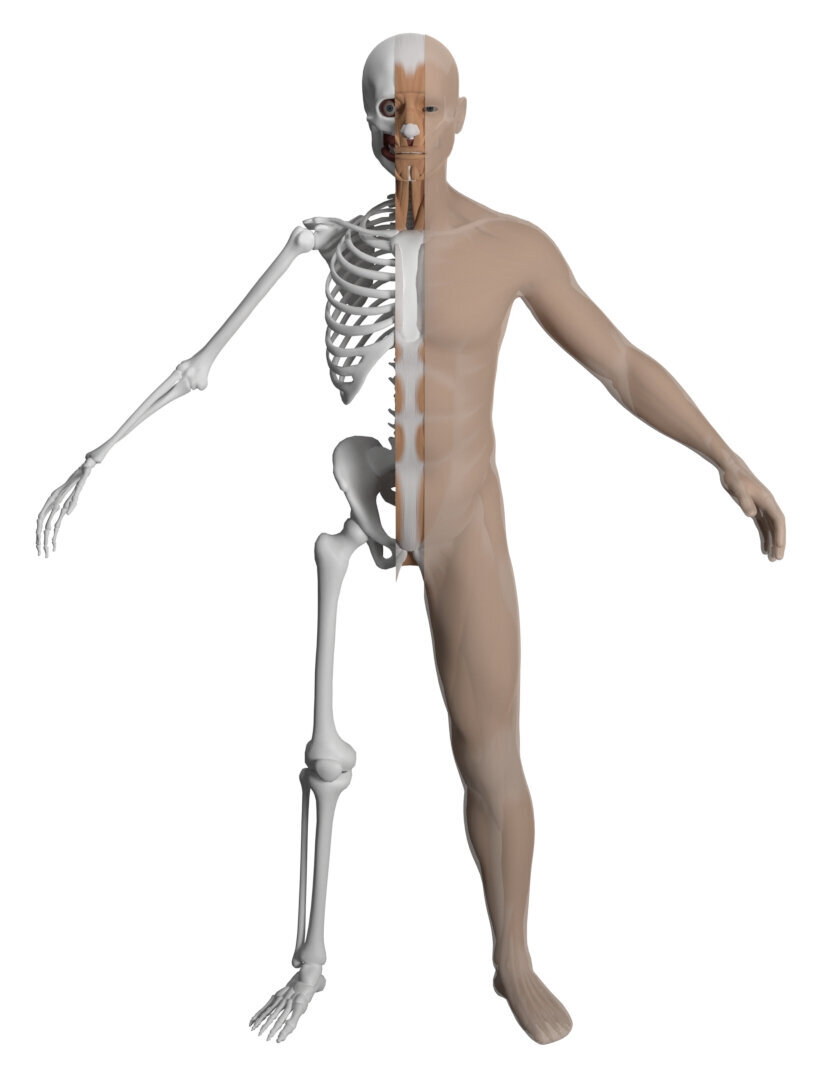}
    \caption{Male and female template model. In this work, we derive an additional skeleton layer that wraps the high resolution skeleton mesh and shares the triangulation with the skin layer.}
    \label{fig:template-model}
\end{figure}

Existing anatomical models, as provided for example by Zygote \cite{zygote} or 3D Scanstore \cite{3dscanstore}, are only available with prohibitive licensing. 
In order to make our model publicly available, we commissioned a 3D artist to build an anatomical template model. 
It provides a male and female template, both including meshes for skin, eyes, mouth, teeth, muscle, and skeleton (\Fig{template-model}). 
All meshes in the male template are consistently topologized with their counterparts in the female template.
With $23752$ vertices, our skin mesh has approximately $3.5$ times more vertices than the popular SMPL \cite{Loper2015SMPL} or STAR \cite{Osman2020STAR} models, allowing us to more accurately model skin geometry (cf.~\Fig{osso_comparison}). 
We follow the \emph{layered model} approach and generate a skeleton wrap that envelopes the high-detail skeleton mesh and has the same triangulation as the skin layer\cite{komaritzan2021inside}. 
This provides a trivial correspondence between skin and skeletal layers.

Our skin surface input data is derived from the European subset of the \emph{CAESAR} database \cite{robinette2002caesar}, consisting of about 1700 3D scans annotated with 3D landmarks and anthropometric measurements. 
To bring all scans into uniform topology and pose, we employ the template fitting approach proposed by Achenbach et al.~\cite{achenbach17}, adapted to use the skin surface of our template model.
This leaves us with $776$ male and $919$ female skin meshes denoted by $\skin_i$.
In the following, all computations are done on the male and female data set separately, due to the anatomical differences especially in the hip and shoulder region. 

From the fitted skin meshes, we use an author-provided implementation of InsideHumans~\cite{komaritzan2021inside} to estimate skeleton layers $\bone_i$, resulting in non-intersecting pairs of skeleton and skin meshes $(\bone_i, \skin_i)$. 
Since the InsideHumans approach excludes the head, hands, and feet region from the skeleton layer, we inherit this limitation.
We denote the set of vertices belonging to these regions by $\excluded$.
Equipped with this data, we can now enlarge our training data set by computing the Cartesian product of skeleton shape and soft tissue distribution in a physically plausible way.

\subsection{Volumetric Deformation Transfer}
\label{sec:vol_deftrans}

We train our model on the Cartesian product of two shape dimensions: skeleton shape and soft tissue distribution. To this end, we transfer the soft tissue of subject $i$ onto the skeleton of subject $j$, which we achieve through deformation transfer \cite{sumner2004, botsch2006deftrans}.

\begin{figure*}
    \centering
    \includegraphics[width=0.245\linewidth]{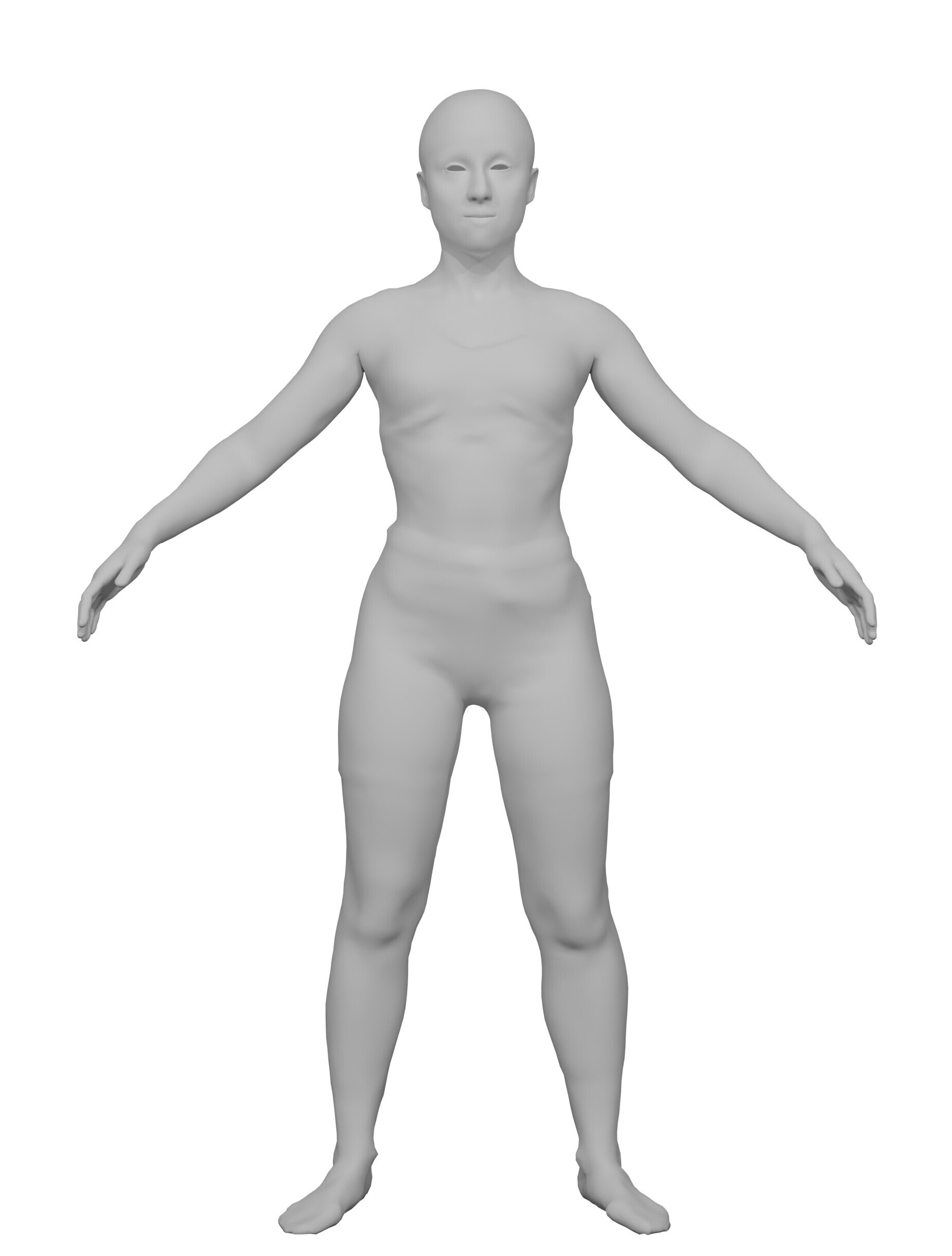}
    \includegraphics[width=0.245\linewidth]{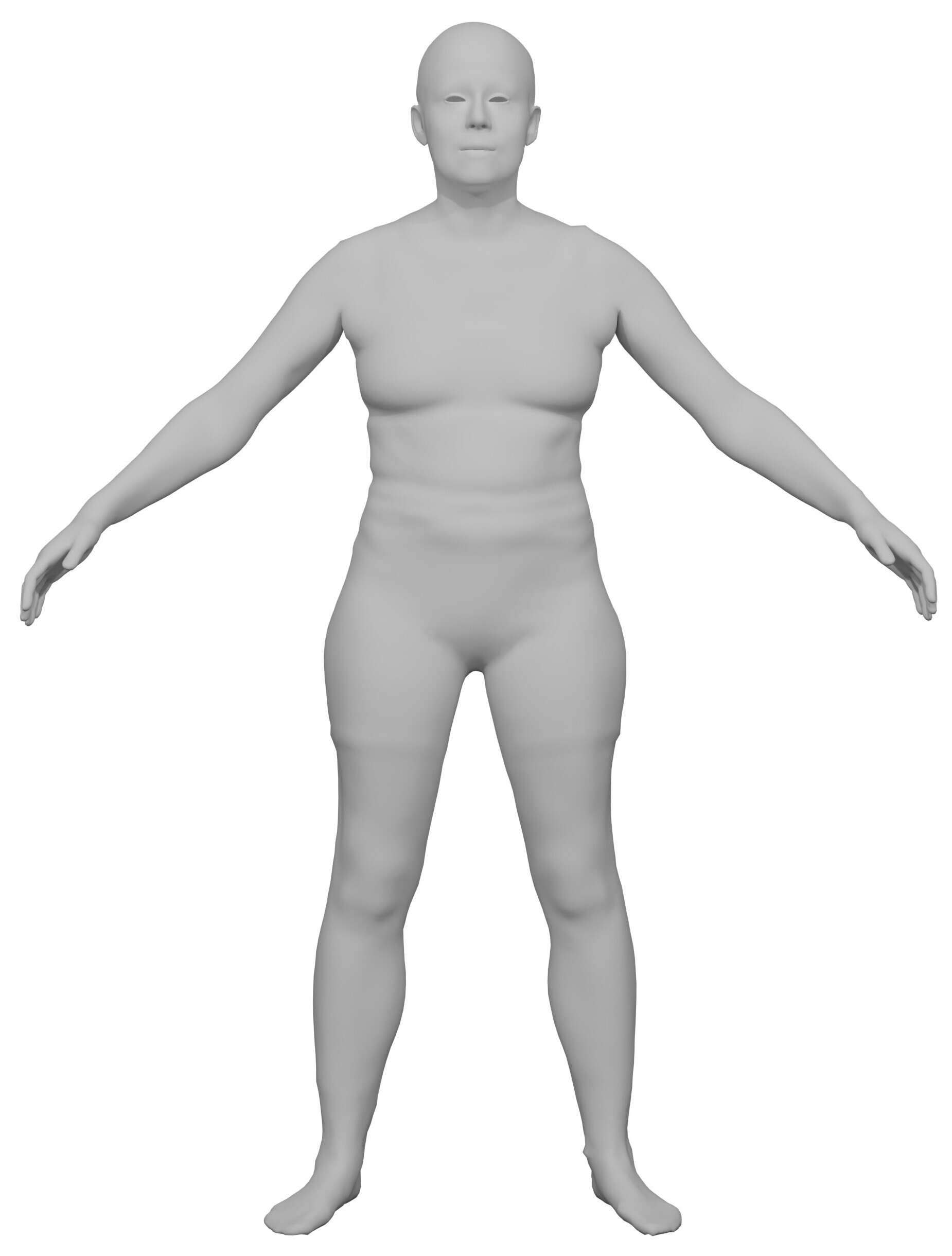}
    \includegraphics[width=0.245\linewidth]{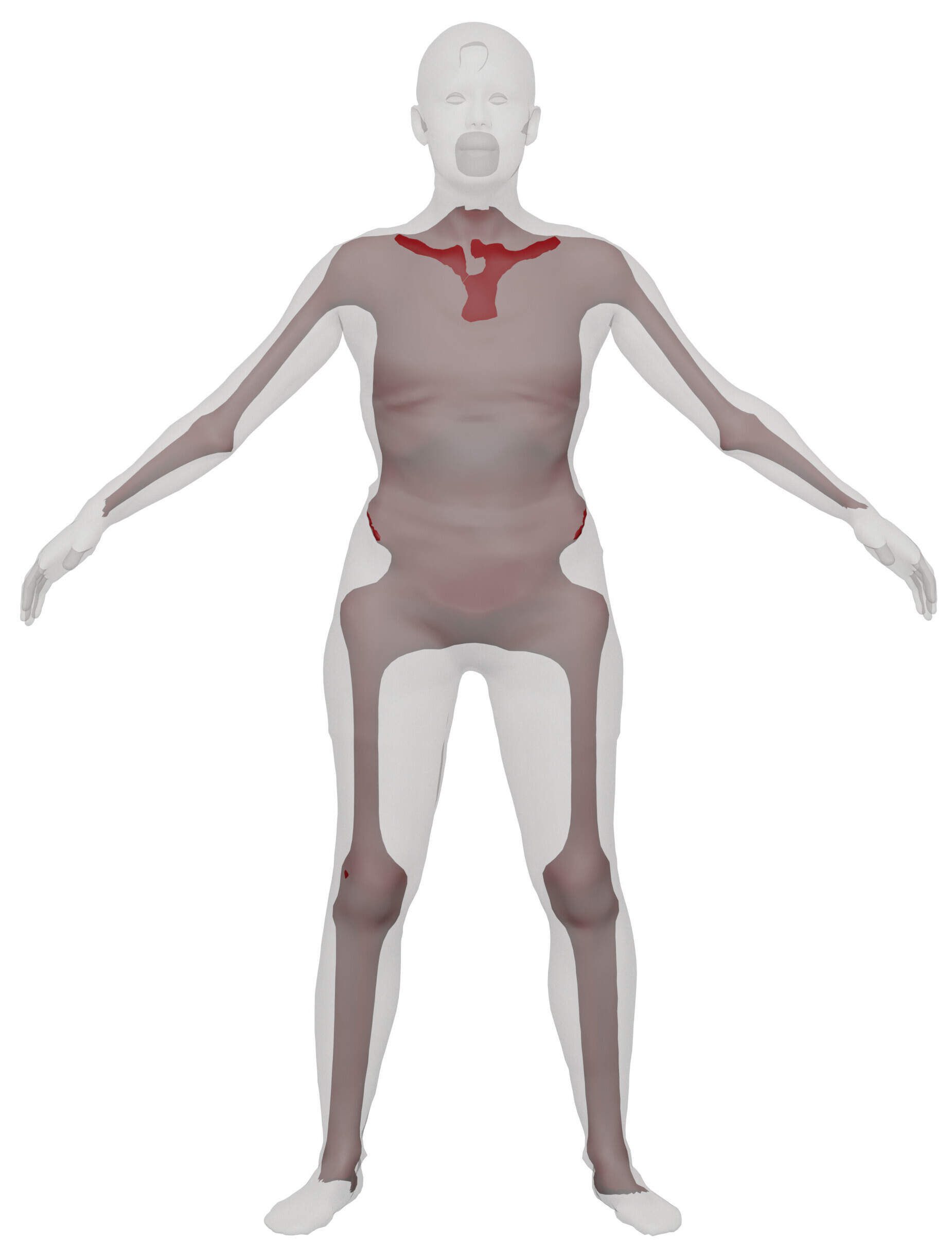}
    \includegraphics[width=0.245\linewidth]{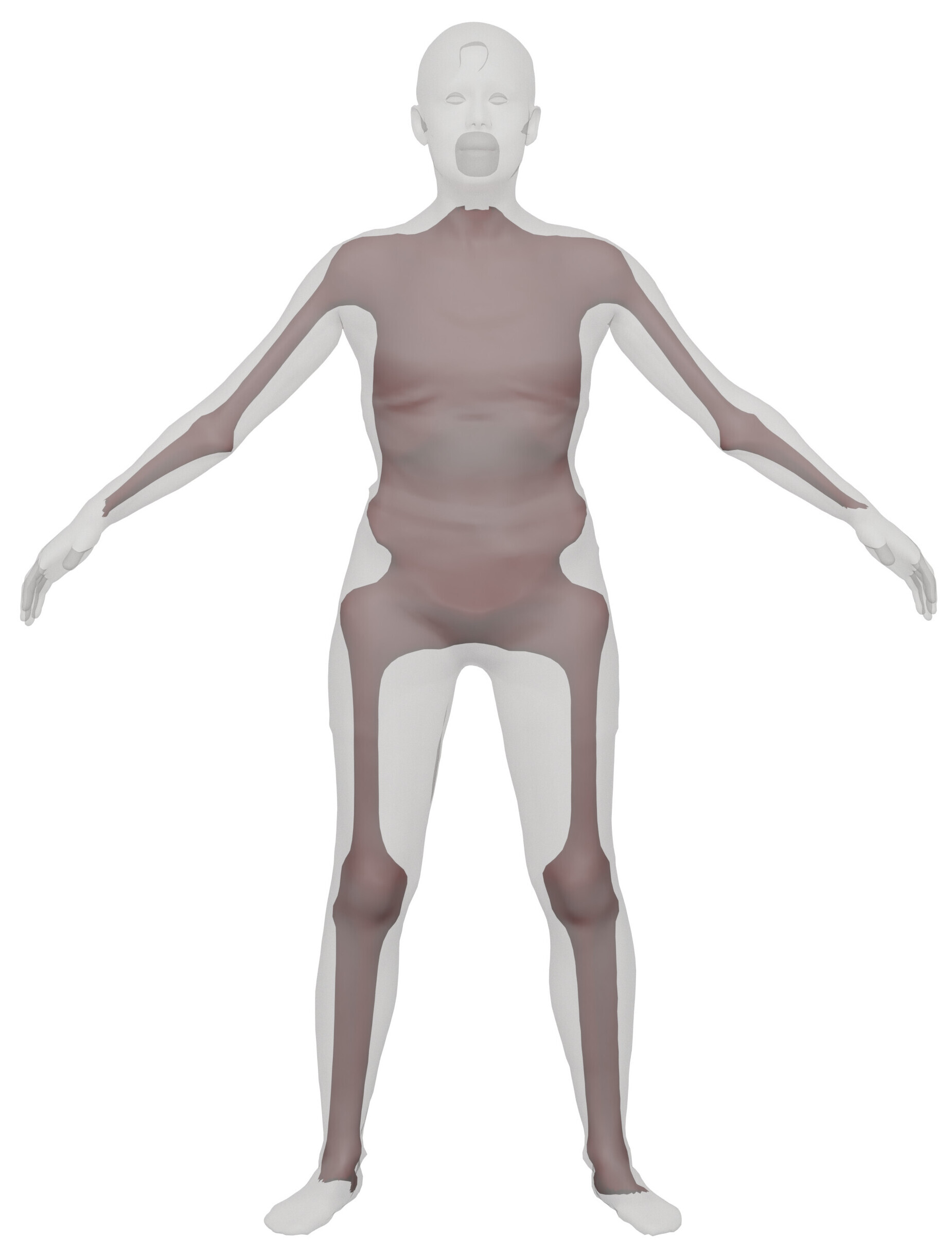}
    \caption{Transferring the soft tissue of the left model onto the skeleton of the center-left model using \emph{surface-based} deformation transfer, the skeleton wrap protrudes the skin (center-right). Our volumetric deformation transfer successfully avoids these artifacts (right).}
    \label{fig:surface_vs_volumetric_deftrans}
\end{figure*}

In the standard formulation of deformation transfer, the deformation gradients are computed from a triangle on $\bone_i$ to the corresponding triangle on $\skin_i$. 
These deformations are then applied to $\bone_j$ in order to generate $\skin_{j}$.
However, as seen in \Fig{surface_vs_volumetric_deftrans} (center right), this formulation can lead to interpenetrations of skin and skeleton. 
These artifacts can happen because the triangle-based deformation gradients between $\bone_i$ and $\skin_i$ do not encode any volumetric information of the soft tissue enclosed in between these two surfaces.
To alleviate this problem we formulate the soft tissue transfer as a \emph{volumetric} deformation transfer problem.

First, we compute the mean skeleton $\bone_\mu$ and mean skin mesh $\skin_\mu$ over all training models. 
Since the two layers share the same triangulation, the corresponding faces between the skeleton and skin layer span prismatic elements that can trivially be split into three tetrahedra. We denote the resulting tetrahedral mesh enclosed between $\bone_\mu$ and $\skin_\mu$ (hence representing the mean soft tissue distribution) as $\tetmesh{S}_\mu$. The vector $\vec{X}_\mu$ containing the stacked vertex positions of $\tetmesh{S}_\mu$ is composed from the vertex positions of the bone mesh $\bone_\mu$ and the skin mesh $\skin_\mu$, denoted by $\vec{X}^\bone_\mu$ and $\vec{X}^\skin_\mu$, respectively.

Transferring the soft tissue layer of subject $i$ onto the skeleton of subject $j$ can then be formulated as a volumetric deformation transfer. The deformation gradients $\mat{F}^t \in \R^{3 \times 3}$ per tetrahedron $t$ encode the deformation from the mean tetrahedral mesh $\tetmesh{S}_\mu$ to the tetrahedral mesh $\tetmesh{S}_i$ of subject $i$.
From the four vertex positions $\vec{x}_1, \vec{x}_2, \vec{x}_3, \vec{x}_4$ of tetrahedron $t$ in $\tetmesh{S}_i$ we build the edge matrix 
\[
    \mat{E}^t_i = \matrix{\vec{x}_1 - \vec{x}_4 ,\, \vec{x}_2 - \vec{x}_4 ,\, \vec{x}_3 - \vec{x}_4}.
\]
The matrix $\mat{E}^t_\mu$ is built analogously from the vertices of $\tetmesh{S}_\mu$. The deformation gradient of tetrahedron $t$ could then be computed as $\mat{F}^t = \mat{E}^t_i \left(\mat{E}^t_\mu\right)^{-1}$. 
However, part of the desired deformation is already explained by the deformation of $\bone_\mu$ to $\bone_j$. To account for this, we express the deformation gradients relative to reference frames on $\bone_\mu$ and $\bone_j$.
Each tetrahedron $t$ can be associated with a triangular face on the skeleton layer $\bone_\mu$ and $\bone_j$, respectively. These triangles define orthonormal reference frames $\mat{R}^t_\mu$ and $\mat{R}^t_j$, respectively, which leads to the final formulation for deformation gradients: 
\begin{equation}
    \mat{F}^t = 
    \mat{R}^t_j \left( \mat{R}^t_\mu \right)\T 
    \mat{E}^t_i \left( \mat{E}^t_\mu \right)^{-1}.
\end{equation}

We then solve for vertex positions $\mat{X}_j$ conforming to these deformation gradients in a least squares sense, while keeping the vertices of the skeleton layer $\bone_j$ and $\excluded$ fixed. Formally, we solve the gradient-based mesh deformation system
\begin{equation}
\label{eq:deformation_transfer_lse}
    \left( \mat{G}\T\mat{D} \, \mat{G} \right) \mat{X}_j 
    \;=\; 
    \left( \mat{G}\T\mat{D} \right) \mat{F},
\end{equation}
with Dirichlet boundary constraints for every vertex belonging to $\bone_j \cup \excluded$. The matrices $\mat{G}\T\mat{D}$ and $\mat{G}$ represent the discrete divergence and gradient operators for tetrahedral meshes~\cite{botsch2006deftrans}, and $\mat{F}$ vertically stacks the desired deformation gradients $\mat{F}^t$. Solving this linear system yields new skin vertices $\mat{X}_j^\skin$.

In order to smoothly blend into the boundary region $\excluded$, we define per-tetrahedron interpolation weights $w^t \in [0,1]$, which decrease based on the distance to $\excluded$.
We use $w^t$ to linearly interpolate between the desired deformation gradients $\mat{F}^t$ and the deformation gradients computed from $\tetmesh{S}_\mu$ to the target subject $\tetmesh{S}_j$, thereby ensuring a smooth transition into $\excluded$. 

\begin{figure*}
    \centering
    \includegraphics[width=\linewidth]{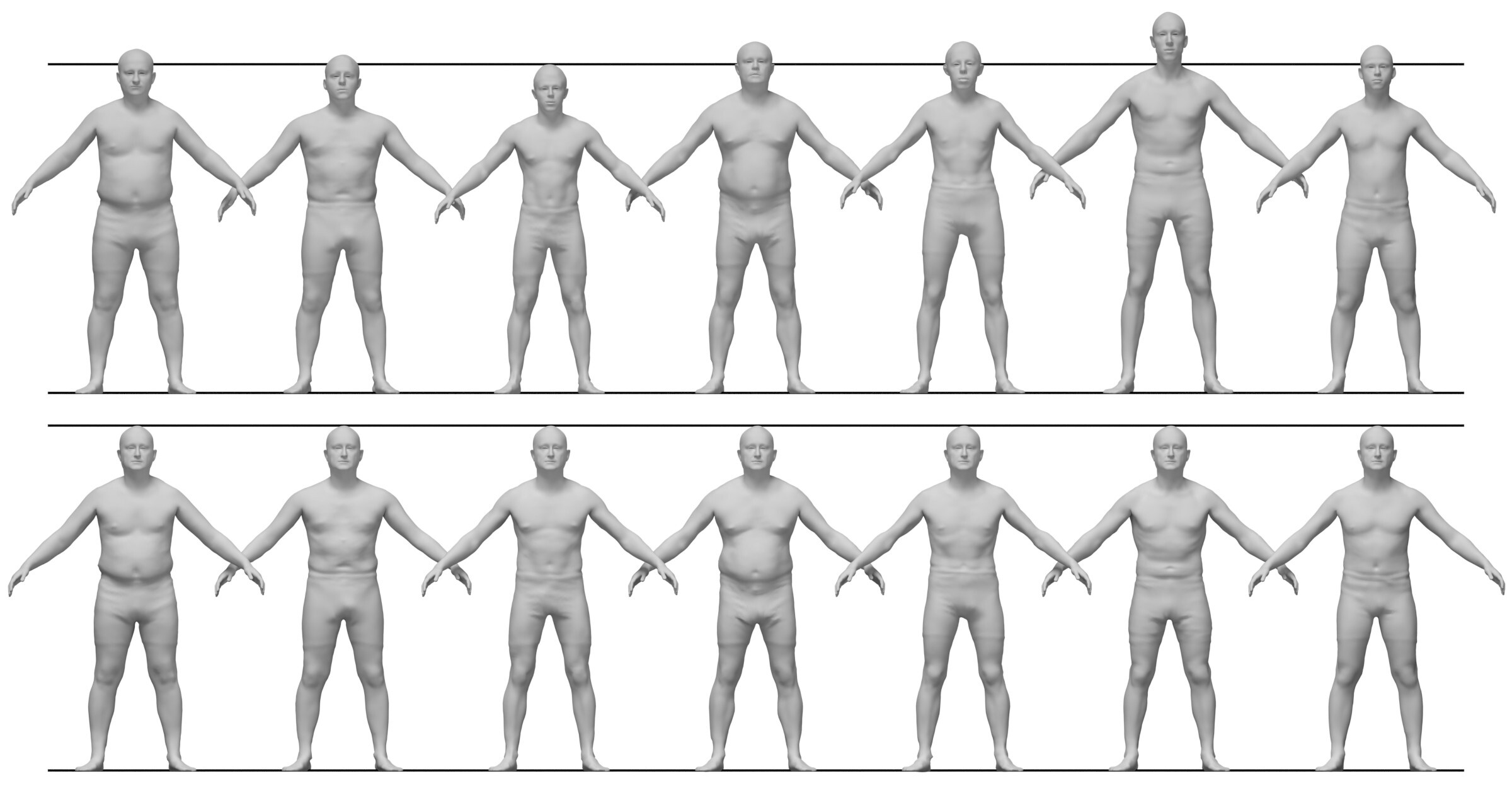}
    \caption{Exemplary results of transferring the soft tissue of various people (top row) onto a single target skeleton via volumetric deformation transfer (bottom row). Note that soft tissue characteristics of the top row and skeletal dimensions of the bottom row are faithfully preserved. }
    \label{fig:volumetric-deformation-transfer}
\end{figure*}

In the presented volumetric formulation, the deformation gradients $\mat{F}^t$ include information about the volumetric stretching and compression of the tetrahedron $t$ of the soft tissue layer. 
Since $\tetmesh{S}_\mu$ and $\tetmesh{S}_i$ do not exhibit inverted elements, the deformation gradients $\mat{F}^t$ do not contain any inversions. 
As such, solving \Eq{deformation_transfer_lse} avoids self-intersections between skin and skeleton (shown in \Fig{surface_vs_volumetric_deftrans}, right), since those would require tetrahedra to invert, which in turn would lead to a high deviation from the target deformation gradient.
\Fig{volumetric-deformation-transfer} shows several examples of transferring the soft tissue distribution of a set of subjects with different height and weight characteristics onto the same skeleton.

\section{Model Learning}
\label{sec:learning}

Our objective is to learn a compact representation of human body shapes. We do so using a specific autoencoder architecture. 
To enable guided and localized shape manipulation, we inject anthropometric measurements into the autoencoder's latent representation.
We measure the length of the torso, arms, and legs on the skeleton, and the circumference of chest, waist, abdomen, and hips on the skin meshes.
By injecting normalized values of those measurements into our latent representation, we form an expressive latent code. 
Our neural network architecture is a convolutional autoencoder with local mesh convolutions based on SpiralNet++ \cite{gong2019spiralnetplus}. Decoupling the two shape dimensions is accomplished by splitting the latent code into two parameter sets: one for skeleton shape and one for soft tissue distribution (\Sec{network-architecture}).
We define a loss function based on the Barlow Twins method \cite{zbontar2021barlow}, which allows us to reduce the redundancy in the latent code (\Sec{training_loss}).

\subsection{Network architecture}
\label{sec:network-architecture}

Our shape compression task is implemented using a convolutional autoencoder. 
To achieve decoupling of skeleton shape and soft tissue distribution, we encode all samples using the encoder of our network, and split the resulting embeddings $\vec{z}$ into two parts: $\vec{z}^{\pbone}$ representing the skeleton and $\vec{z}^{\pskin}$ representing the soft tissue distribution. 
To facilitate semantic control in the latent space, the normalized values of the measurements taken on the original meshes are then appended to one of the two parts of the latent representations. 
Measurements taken on the skeleton are appended to $\vec{z}^{\pbone}$, skin measurements are appended to $\vec{z}^{\pskin}$.

\begin{figure}
    \centering
    \includegraphics[width=0.58\linewidth]{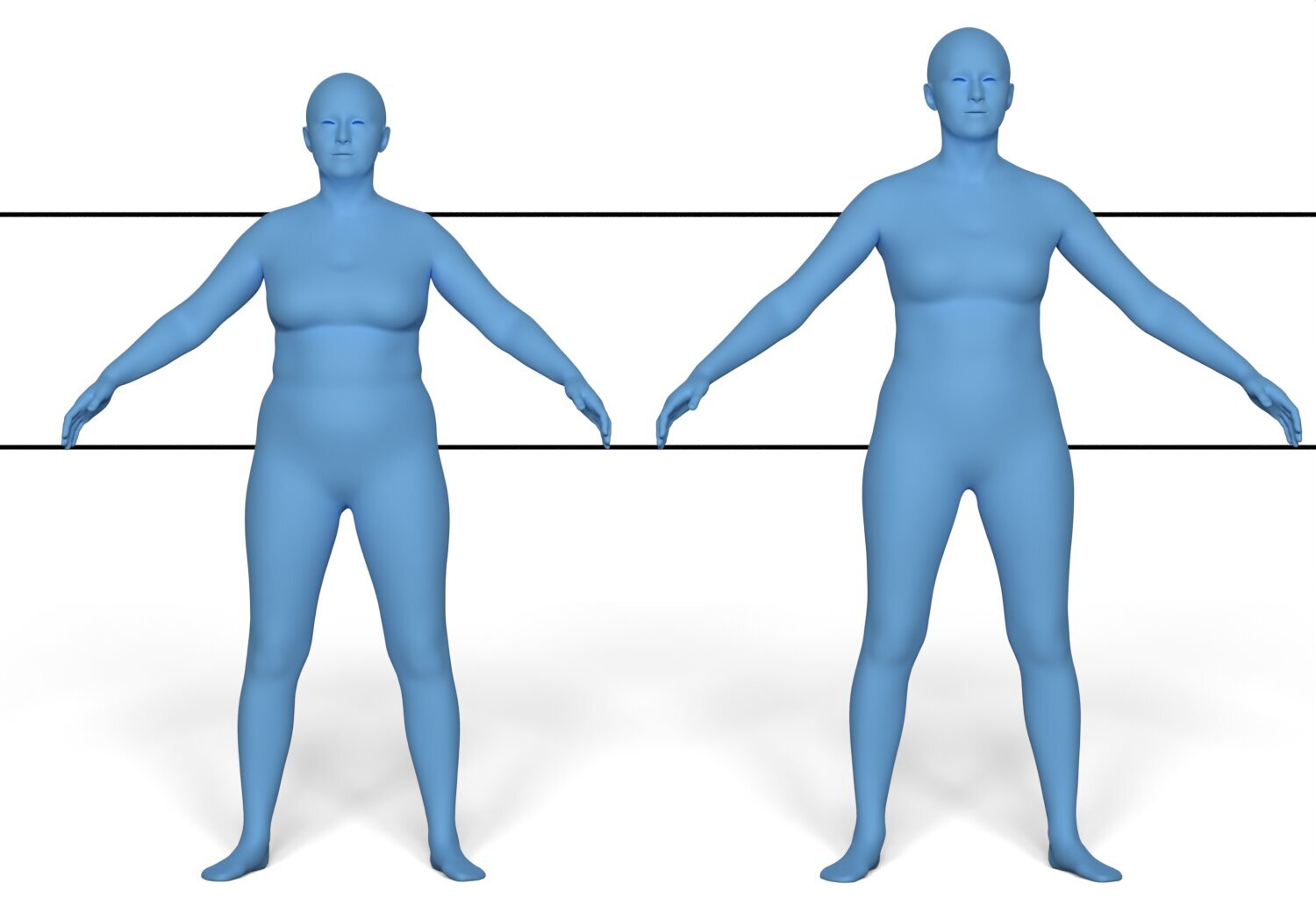}
    \caption{Representing skeletons and skins in PCA subspaces separates their parameters, but the global nature of PCA prevents localized changes: Increasing the arm length of the left model also causes the body height to increase (right).}
    \label{fig:bt-pca-model-arms}
\end{figure}

As a first design for the shape compression task, we experimented with utilizing two separate PCA models for the skeleton and soft tissue distribution. The PCA weights then formed the input to our autoencoder, which learned to decouple the two shape dimensions.
This is in line with the OSSO approach \cite{Keller2022OSSO}, where the correlation between skin and skeleton shape is learned by a linear regressor between two PCA subspaces. 
We found that the resulting model separates the skeleton and soft tissue distribution, but only provides global shape control when modifying semantic parameters in the latent space, due to the global influence of the PCA weights.
\Fig{bt-pca-model-arms} shows an example of the global influence, where modifying arm length also changes the body height. 

\begin{figure*}
    \centering
    \includegraphics[width=\linewidth]{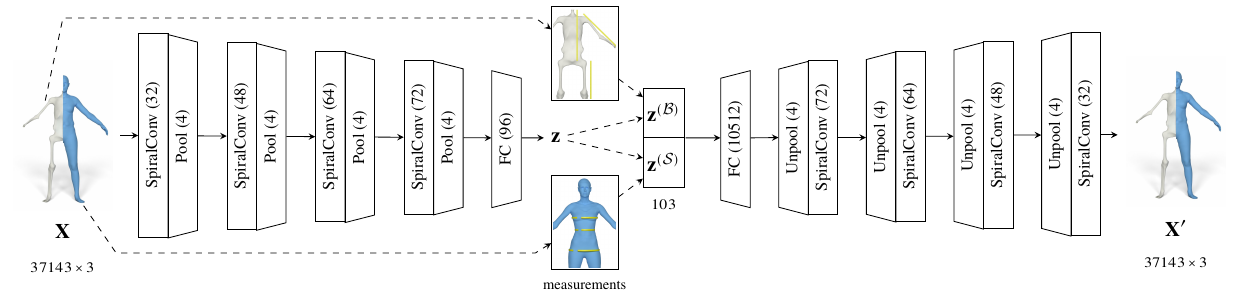}
    \caption{Our network architecture is based on four SpiralNet++\cite{gong2019spiralnetplus} convolution and pooling blocks in the encoder. A final dense layer is connecting the last layer of the encoder to achieve our embeddings $\vec{z}$. We divide the embeddings into two parts, one for the skeleton and the other one for the soft tissue distribution. We append the normalized values of the measurements (the lengths of torso, arms and legs and the circumferences of chest, waist, abdomen, and hips) taken on the input mesh via a skip connection to the latent code. The decoder is the reversed order of the encoder using four unpooling and convolution blocks.}
    \label{fig:spiral-net-architecture}
\end{figure*}

To mitigate the global effects, we opt for an autoencoder using the SpiralNet++ approach \cite{gong2019spiralnetplus}, which utilizes a mesh convolution and pooling operator. This design enables local shape control when modifying the entries in the latent space that correspond to the anthropometric measurements.

The structure of our autoencoder is shown in \Fig{spiral-net-architecture}. 
The samples $\mat{X} \in \R^{37143\times3}$ drawn from our Cartesian product data set (\Sec{vol_deftrans}) consist of 
the vertex positions $\mat{X}^\skin$ and $\mat{X}^\bone$ of the skin mesh and the skeleton wrap (the latter excluding vertices in $\excluded$ belonging to head, hands, and feet). 
We normalize the vertex positions before using them as input for our autoencoder.
Our latent code utilizes $48$ parameters for the skeleton and soft tissue distribution each.
We take three measurements on the skeleton (torso, arm and leg length), four measurements on the skin (chest, waist, abdomen and hip circumference), and append them to the resulting embeddings via skip connections.
This results in a total of $103$ parameters in the latent space.

\subsection{Cross-correlation Loss}
\label{sec:training_loss}

Our encoder creates a latent representation $\vec{z}$ for each sample $\mat{X}$ in the Cartesian data set. 
Samples with identical skeleton shape should result in identical skeleton embeddings $\vec{z}^{\pbone}$, while samples that share soft tissue distribution should result in identical soft tissue embeddings $\vec{z}^{\pskin}$.
We achieve this using a self-supervised learning approach based on Barlow Twins \cite{zbontar2021barlow}, where the loss formulation penalizes dissimilar embeddings for similar input samples.

\begin{figure}
    \centering
    \includegraphics[width=0.58\linewidth]{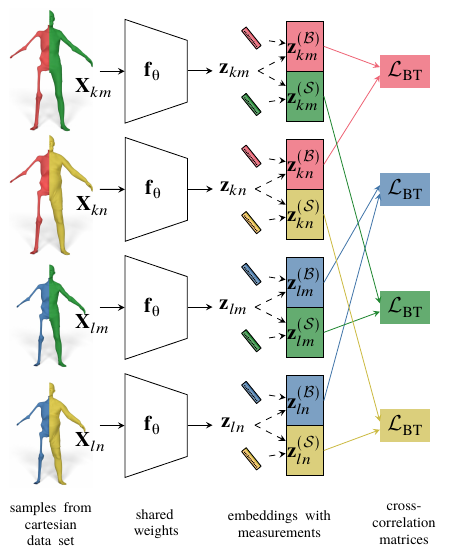}
    \caption{
    Processing a quadruplet of samples in the encoder and dividing the embeddings into two parts: one for the skeleton and one for the soft tissue distribution. After separation the normalized measurements are appended to the respective embeddings.
    If the divided embeddings are noted the same way, they are merged for the entire batch. In order to calculate the cross-correlation loss between pairs, the cross-correlation matrices are calculated for the positions colored in the same way.}
    \label{fig:quadruplet-processing}
\end{figure}

We extend the concept of pairs in Barlow Twins by using quadruplets, built by all four combinations of skeletal and soft tissue distribution from two samples each for pairs of skeleton and skin meshes. 
We randomly select two different indices $k, l$ for skeletons and two different indices for the distribution of soft tissues $m, n$ from our training data set. 
Let $\mat{X}_{km}$ denote the vertex positions resulting from transferring the soft tissue distribution of subject $m$ onto the skeleton of subject $k$ via volumetric deformation  transfer (\Sec{vol_deftrans}).
From the chosen indices $\left(k, l, m, n\right)$ we create a quadruplet containing the entries $\left(\mat{X}_{km}, \mat{X}_{kn}, \mat{X}_{lm}, \mat{X}_{ln}\right)$, such that each entry shares either its skeleton or its soft tissue distribution with two of the other entries.
These quadruplets are processed in batches by our autoencoder. The forward process of the encoder for one quadruplet in a batch is visualized in \Fig{quadruplet-processing}.

Following the Barlow Twins method\cite{zbontar2021barlow}, we reduce the redundancy in the embeddings of common features -- resulting from samples that share either the skeleton shape or the soft tissue distribution -- by computing empirical cross-correlation matrices of the embeddings and penalizing their deviation from the identity matrix. 
The cross correlation loss is defined as  
\begin{equation}
    \label{eq:bt_loss}
    \mathcal{L}_\text{BT} = \sum_{i} \left(1 - C_{ii}\right)^2 + \lambda \sum_{i} \sum_{j \neq i} C_{ij}^2,
\end{equation}
with 
\begin{equation}
    \label{eq:cross_correlations_matrices}
    C_{ij} = \sum_b z_{b,i}^{A} \cdot z_{b,j}^{B}. 
\end{equation}
The batch dimension is denoted by $b$, while the index dimensions of the network output are represented by $i$ and $j$. 
$\lambda$ is a trainable hyperparameter to weigh the importance of off-diagonal entries being close to $0$ in the empirical cross-correlation matrices. 
$\vec{z}^{(A)}$ and $\vec{z}^{(B)}$ are batches of embeddings, which are selected as described in the following.
Note that \Eq{cross_correlations_matrices} differs from the original definition in that we do not use batch normalization on the embeddings before calculating the entries of the cross-correlation matrices $C_{ij}$.
Our model achieves greater accuracy without batch normalization and enables more efficient manual modifications to the reconstructed meshes.

We rearrange the embeddings in a batch to group all components with similarities on the source data set. These embeddings should have a minimal redundancy when originating from the same distribution. This is indicated when sharing one of their indices in the training data set $k$, $l$, $m$, or $n$. 
We can form four cross-correlation matrices for the quadruplets in the batch: 
$\vec{z}_{km}^{\pbone} \otimes {\vec{z}_{kn}^{\pbone}}$, 
$\vec{z}_{lm}^{\pbone} \otimes {\vec{z}_{ln}^{\pbone}}$, 
$\vec{z}_{km}^{\pskin} \otimes {\vec{z}_{lm}^{\pskin}}$,
and $\vec{z}_{kn}^{\pskin} \otimes {\vec{z}_{ln}^{\pskin}}$, 
where $\otimes$ denotes the outer product of the batched embeddings.
We calculate the loss using \Eq{bt_loss} and sum up the total loss for all four matrices of the batch to get our redundancy loss $\mathcal{L}_Q$. Note that we do not need to minimize the redundancy between skeleton and soft tissue parameters directly to learn a separation.

Let $\mathcal{L}_R$ denote the $L^1$ reconstruction loss over all samples of the quadruplets in the batch. 
We train our network to minimize the combined loss function 
\begin{equation}
    \label{eq:combined_loss}
    \mathcal{L} = \mathcal{L}_R + \beta \mathcal{L}_Q, 
\end{equation}
where $\beta$ is a trainable hyperparameter that balances the importance of reconstruction and redundancy reduction in our loss function. 

We train the autoencoder using the Adam Optimizer \cite{kingma2015Adam}. 
A randomized hyperparameter search is conducted and the highest performing model in the validation data set was selected. 
We trained our models over the complete training set, resulting in $12$ epochs for female and $27$ for males. This model was trained using a learning rate of $1.71 \cdot 10^{-4}$ for females and $1.78 \cdot 10^{-4}$ for male. 
We used redundancy importance values with $\beta = 0.52$ for females and $\beta = 0.42$ for males. The optimal importance hyperparameter for off-diagonal entries to achieve best performance was $\lambda = 2.3\cdot 10^{-2}$ for females and $\lambda = 4.3 \cdot 10^{-2}$ for males.

\section{Post-Processing}
\label{sec:postprocessing}

After the inference of the decoder, the resulting meshes might show certain artifacts. We observed asymmetries in the face region, which are amplified when modifying the latent code towards the boundary of the learned distribution.
This is in part due to the fact that the variance of the face region was not part of the modification in the training data set (\Sec{template_and_caesar}). 
To mitigate potential artifacts, we apply three post-processing steps after decoding: (i) the face region is symmetrized, (ii) the resulting skin surface is smoothed and (iii) intersections between the skeleton and skin layer are resolved.
As a final step, we embed the high-resolution anatomical skeleton into the skeleton wrap using a triharmonic space warp.

\subsection{Face Symmetrizing \& Smoothing}

After the inference of the decoder, we approximately symmetrize the face region by adapting the approach of Mitra et al.~\cite{mitra2007symmetrization}.
A reflective symmetry plane is defined at the center of the head, based on which corresponding vertex pairs $(v_i, v_j)$ on both sides can be determined. The $y$-coordinate of these vertex pairs are adjusted to match approximately: $y_{i}' = \frac{3}{4} y_i + \frac{1}{4} y_j$. 
Afterwards, one explicit smoothing step \cite{desbrun1999fairing} is performed on the skin mesh in order to reduce high frequency noise, which may occur when applying drastic changes to the latent parameters. 

\subsection{Intersection Avoidance}
\label{sec:intersection_avoidance}

After decoding from the latent space, the resulting skeleton $\bar{\bone}$ with vertices $\set{V}$ might slightly protrude the skin layer $\skin$, especially when the target skin measurements in the latent code are set to lower values. We detect protruding triangles and add all vertices belonging to its two-ring neighborhood to the collision set $\collisions$.

When inferring the skeleton for a skin $\skin$ given by a 3D scan (as demonstrated in \Sec{modifying_virtual_humans}), we want to keep the vertices on $\skin$ fixed, as they can be considered ground truth. As such, in order to resolve the detected collisions, we solve for a new skeleton layer $\bone$ by minimizing
\begin{equation}
    \label{eq:collision_resolve}
    E \of{\bone} = 
    E_\mathrm{reg} \of{\bone, \bar{\bone}} + 
    E_\mathrm{close} \of{\bone, \bar{\bone}} +
    w_\mathrm{coll} E_\mathrm{coll} \of{\bone, \skin},
\end{equation}
where $E_\mathrm{reg}$ is a bending constraint on the skeleton layer:
\begin{equation}
    \label{eq:collision_bending_constraint}
    E_\mathrm{reg} \of{\bone, \bar{\bone}} = 
    \frac{1}{2} \sum_{\vec{x}_i \in \bone} A_i \norm{\laplace\vec{x}_i - \mat{R}_i\laplace \bar{\vec{x}}_i}^2 .
\end{equation}
$\mat{R}_i \in SO(3)$ denotes the rotation matrix optimally aligning the vertex Laplacians between the resolved surface $\bone$ and the initial surface $\bar{\bone}$. The Laplace operator is discretized using cotangent weights and Voronoi areas $A_i$ \cite{botsch10},

$E_\mathrm{close}$ constrains vertices that are not part of the collision set $\collisions$ to stay close to their original position:
\begin{equation}
    E_\mathrm{close} \of{\bone, \bar{\bone}} = 
    \frac{1}{\abs{\set{V} \setminus \collisions}} \sum_{\vec{x}_i \notin \collisions} \norm{\vec{x}_i - \bar{\vec{x}}_i}^2,
\end{equation}
and $E_\mathrm{coll}$ defines the collision avoidance term: 
\begin{equation}
    \label{eq:collision_avoidance_term}
    E_\mathrm{coll} \of{\bone, \skin} = 
    \frac{1}{\abs{\collisions}} \sum_{\vec{x}_i \in \collisions} w_i \norm{\vec{x}_i - \vec{\pi}_\skin\of{\vec{x}_i}}^2,
\end{equation}
where $\vec{\pi}_\skin\of{\vec{x}_i}$ projects vertex $\vec{x}_i$ to lie \SI{2.5}{\milli\meter} beneath the colliding triangle's plane on the skin $\skin$.

We iteratively minimize \Eq{collision_resolve} via the projective dynamics solver implemented in the ShapeOp library \cite{deuss15}.
We set $w_\mathrm{coll} = 50$, $w_i = 1$, and progressively increase the per-vertex collision avoidance weight $w_i$ by $1$ each iteration in which the collision could not be resolved. 
After each iteration, the Laplacian of the initial state $\bar{\bone}$ in \Eq{collision_bending_constraint} is updated to the current solution, thereby making the skeleton layer slightly less rigid.
Following this optimization scheme, we could reliably resolve all between-layer-collisions in our tests.

Note that when modifying the soft tissue distribution over a given skeleton $\bone$, we analogously keep the vertices on $\bone$ fixed, and solve for a new intersection-free skin layer $\skin$.

\subsection{Embedding High-resolution Skeleton}

Once an intersection-free pair $\left(\bone, \skin\right)$ is generated, we embed the high-resolution skeleton mesh $\highresskel$ by following Komaritzan et al.~\cite{komaritzan2021inside}, i.e., using a space warp based on triharmonic radial basis functions \cite{botsch2005}. 
The matrix of the involved linear system depends on the template skeleton $\template{\bone}$ only and hence can be pre-factorized. After generating a new skeleton $\bone$, the solution can be inferred by back-substitution, and the space warp can efficiently be evaluated to embed the high-resolution anatomical skeleton.

\section{Results and Applications}
The resulting {\scshape TailorMe} model allows local shape manipulation based on the injected measurements in the latent space. 
For a demonstration of the final model, we refer the reader to the accompanying video.
In the following, we evaluate the performance of the model on our test data set, 
and compare our approach to the related approaches of OSSO \cite{Keller2022OSSO}, MLM \cite{achenbach2018} and standard PCA approaches \cite{allen06learning, piryankova2014}.
Finally, we demonstrate the modification of 3D-scanned realistic virtual humans. 

\subsection{Model Evaluation}
\label{sec:model_evaluation}
To quantitatively evaluate the fit of our trained model defined in \Sec{learning}, we do not perform the post-processing decribed in \Sec{postprocessing}.
We use separate data sets for training and model evaluation. The training subset is utilized for model optimization, while a validation subset is used to conduct an automated evaluation and to estimate the models capability for generalization. 
To minimize the validation set bias, we utilize a third subset of our data set for testing. The splitting is done such that the skeleton and soft tissue distribution of an individual ends up in only one of these subsets.

We divide our data set into training, validation, and test such that the number of Cartesian pairs in the final data set corresponds to a ratio of \mbox{$8$:$1$:$1$}.
Let $a$, $b$, and $c$ denote the number of samples in the training, validation, and test set, respectively. We want the ratio of squared subset samples $a^2 : b^2 : c^2$ to match the target split ratio of $8:1:1$.
This results in (539, 190, 190) samples for the female and (456, 160, 160) samples for the male data set.

\begin{figure}
    \centering
    \includegraphics[width=0.58\linewidth]{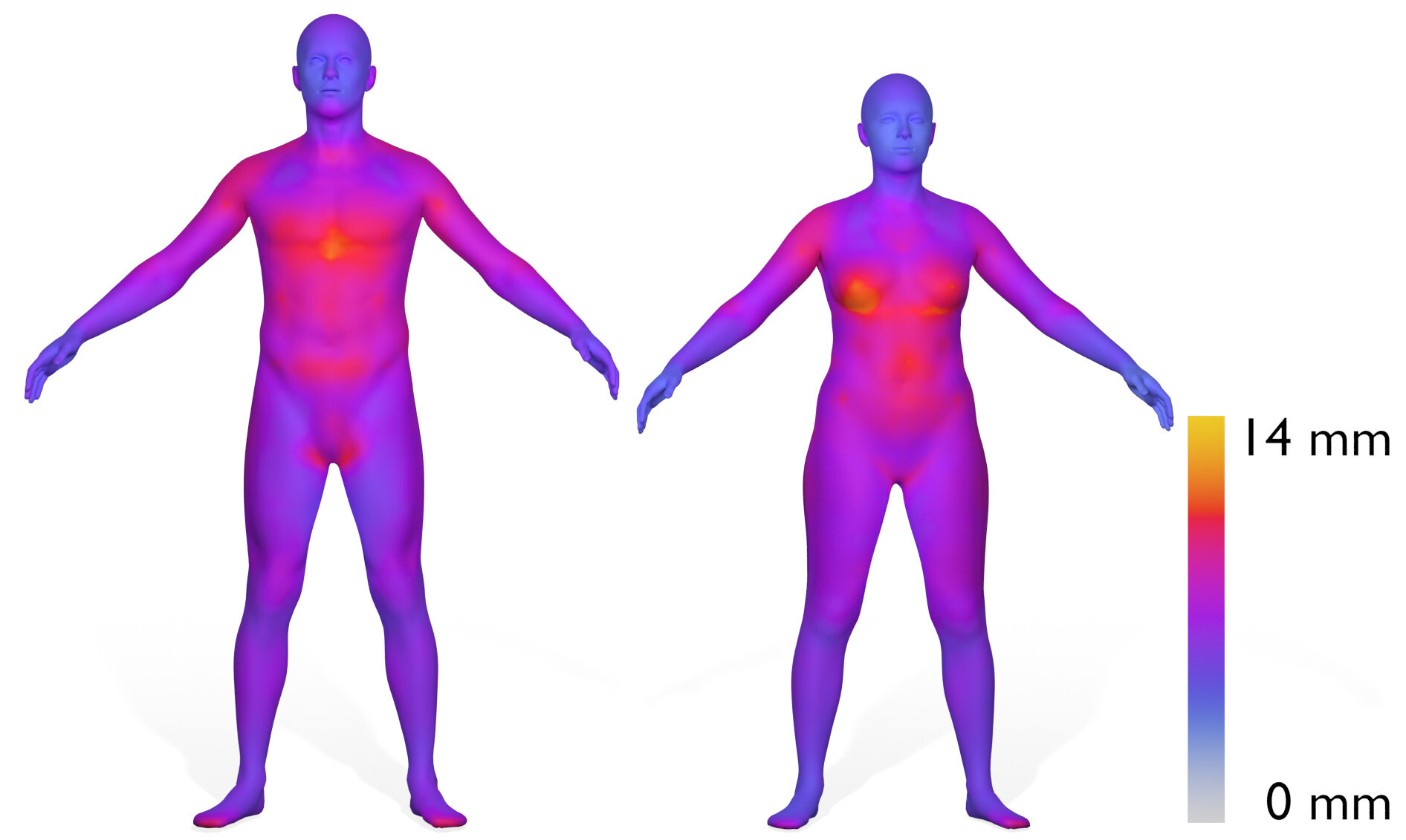}
    \caption{Mean Euclidean vertex distance when evaluating over all samples in the male (left) and female (right) test data set. Our model achieves a maximum Euclidean distance of 11.5~mm for males and 13.9~mm for females.}
    \label{fig:med_fine_model}
\end{figure}

\Fig{med_fine_model} displays our model's reproduction error for the skin, measured as per-vertex distance averaged over all meshes in the test data set, when using the decoder back propagation method.
The vertex distances of the fitted skins are evenly distributed on the limbs and face, but in the chest and abdomen regions the largest average deviation from the target is observed. Overall, our model attains a maximum per-vertex error of \SI{13.9}{\milli\meter} over all samples in the test data set. 

The mean absolute error is the $L^1$ loss for the predicted mesh $\mat{X}'$ to the input mesh $\mat{X}$ with $n$ vertices, which is defined as 
\begin{equation}
    \mathcal{L}_1 = \frac{1}{3n} \norm{ \mat{X} - \mat{X}' }_1 = \frac{1}{3n} \sum_{i=1}^{n} \norm{ \vec{x}_{i} - \vec{x}'_{i}}_1.
\end{equation}

When using the encoder and decoder for reproduction of the test data set, we achieve a mean absolute error for the skeleton wrap and the skin of 
\SI{5.2(1.5)}{\milli\meter} for females and \SI{5.4(1.5)}{\milli\meter} for males.
The cross-correlation matrices in \Eq{cross_correlations_matrices} converge to the identity matrix. The individual mesh measurements positively correlate with each other as the circumferential measurements on the original meshes show a strong connection. 

When inferring a skeleton from a given skin $\vec{X}^\skin$, we use the Adam optimizer \cite{kingma2015Adam} on the latent parameters $\vec{z}$ and minimize the $L^1$ error for the skin arising from decoding $\vec{z}$ to the target skin of the sample.
For skins we achieve a mean absolute reproduction error on the test data set of \SI{2.8(0.5)}{\milli\meter} for females and \SI{2.9(0.5)}{\milli\meter} for males. For skeleton wraps we reach a mean absolute error of \SI{6.3(1.4)}{\milli\meter} for females and \SI{6.9(1.7)}{\milli\meter} for males.



\subsection{Comparison to OSSO}

\begin{figure}
    \centering
    \includegraphics[width=0.28\linewidth]{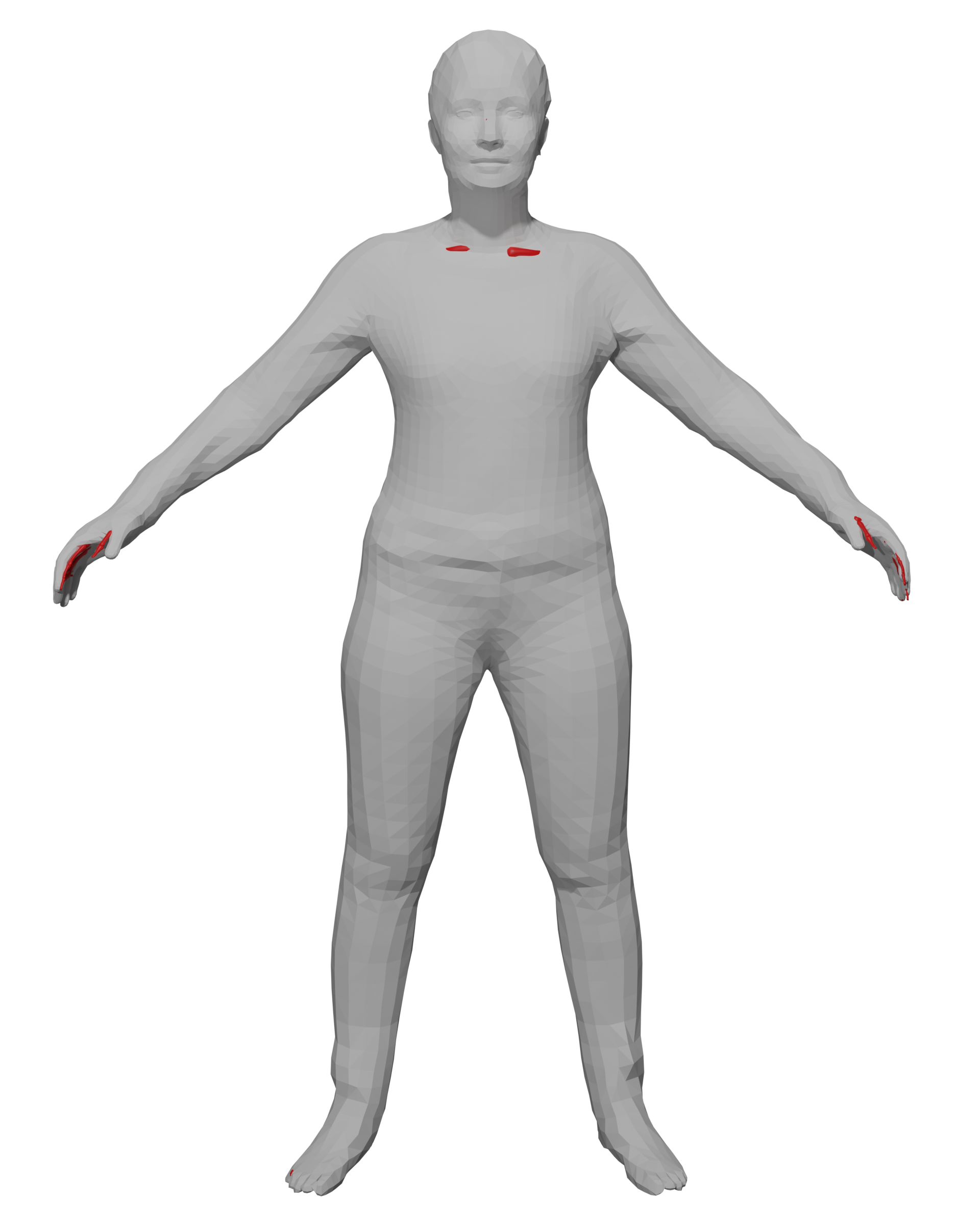}
    \includegraphics[width=0.28\linewidth]{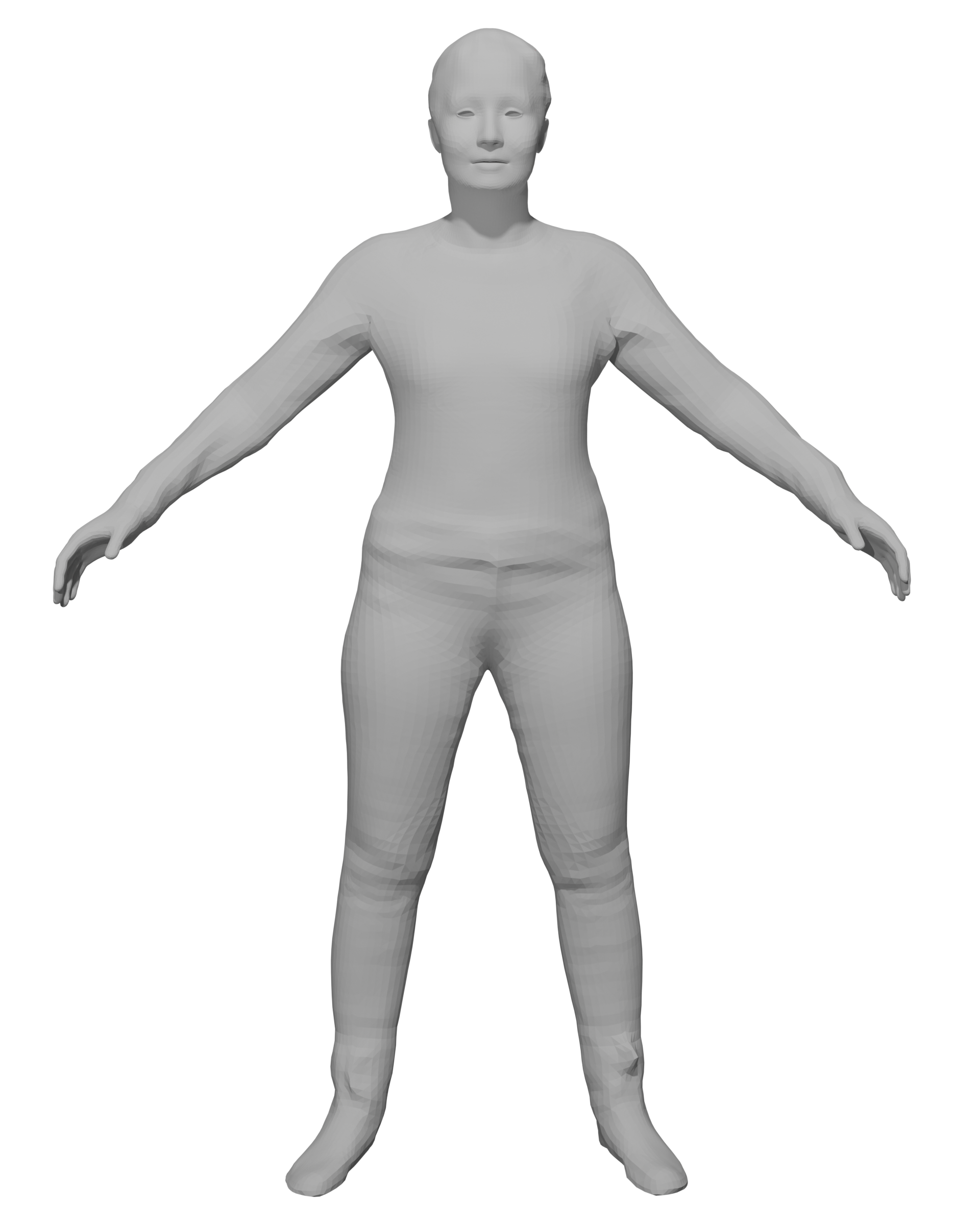}
    \includegraphics[width=0.56\linewidth]{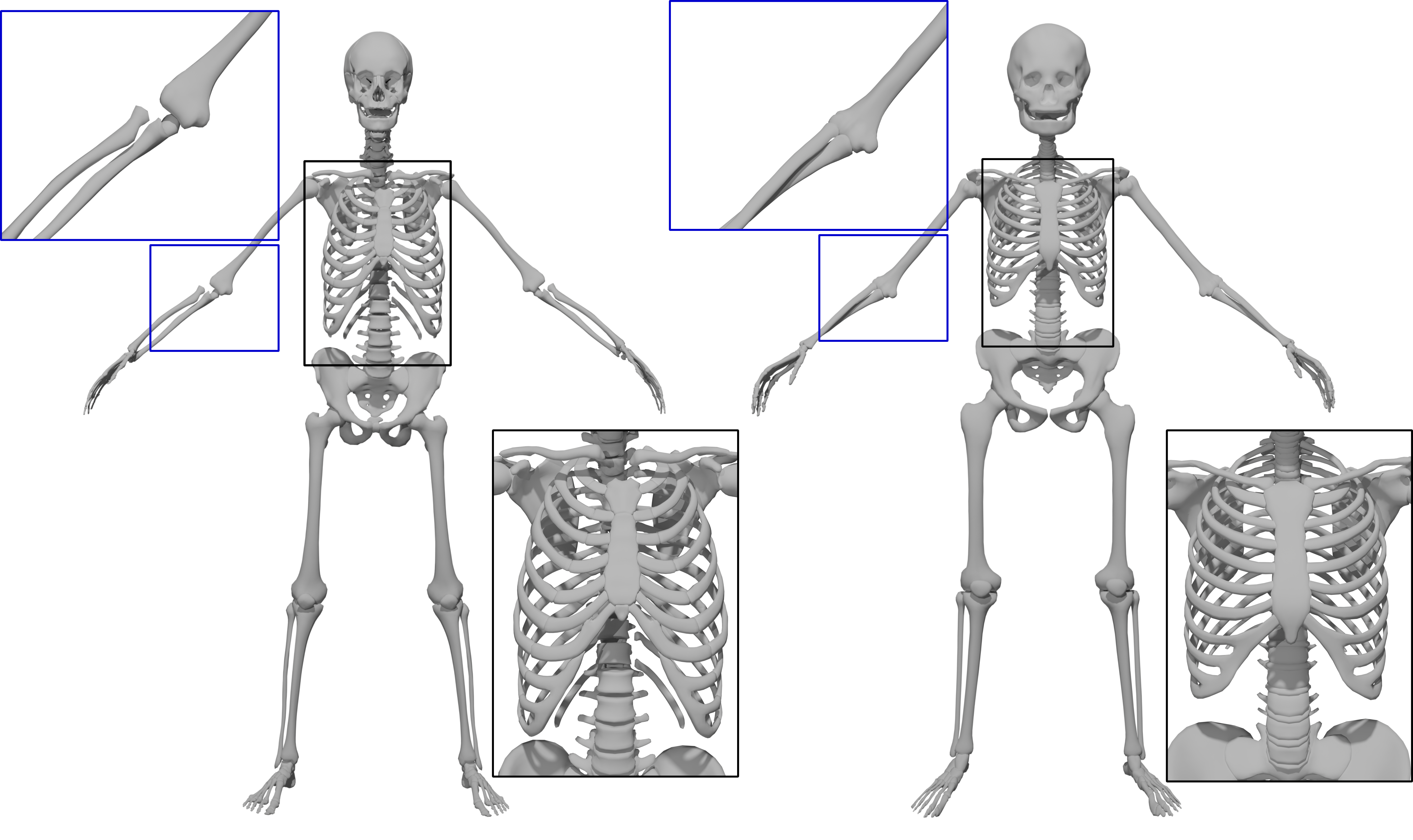}
    \caption{Comparison of OSSO \cite{Keller2022OSSO} (left) with our approach (right). The skeleton OSSO infers protrudes the skin (shown in red), we resolve these kinds of collisions (top row). The rib cage inferred by OSSO is skewed (black inset) and the stitched puppet model results in gaps between bone structures (blue inset) (bottom row).}
    \label{fig:osso_comparison}
\end{figure}

We qualitatively compare our work to the OSSO approach \cite{Keller2022OSSO}. This method computes a linear regressor between skin and skeleton PCA shape spaces, after fitting both shape models to a set of DXA Scans. As DXA scans are taken in a lying pose, OSSO first reposes a given skin mesh to this pose, infers skeleton shape there, and finally reposes the given result to the input pose. As seen in \Fig{osso_comparison}, when compared to our skeleton prediction, the skeleton inferred by OSSO exhibits a skewed and unnaturally shifted rib cage as well as large gaps between bone structures, such as the elbow region or in between ribs and spine. 
Moreover, OSSO's skeleton protrudes through the given skin, while our method resolves these intersections (\Sec{intersection_avoidance}).
In contrast to OSSO, however, our model can only deliver skin and skeleton surfaces in A-pose.

\subsection{Comparison to MLM}

We compare our model to the multilinear model (MLM) presented in~\cite{achenbach2018}. 
The MLM approach requires the computation of a 3D tensor to separate the skeleton and soft tissue dimensions. Given our model with $51$ skeleton and $52$ soft tissue parameters, the MLM requires a total of 292M parameters. Our method requires two orders of magnitude fewer parameters (2.1M) to process the same number of input variables. 
When applying the MLM approach to our training data, we found that the decoupling process of the two parameter sets is incomplete. This effect can also be observed in the original work. 
Achenbach et al. \cite{achenbach2018} provide a demo application at \url{https://ls7-gv.cs.tu-dortmund.de/publications/2018/vcbm18/vcbm18.html}, where changing the first skull parameter causes subsequent changes to the first \emph{soft tissue} parameter to also change the resulting skull shape.  

\subsection{Comparison to Surface PCA}

\begin{figure}
    \centering
    \includegraphics[width=0.58\linewidth]{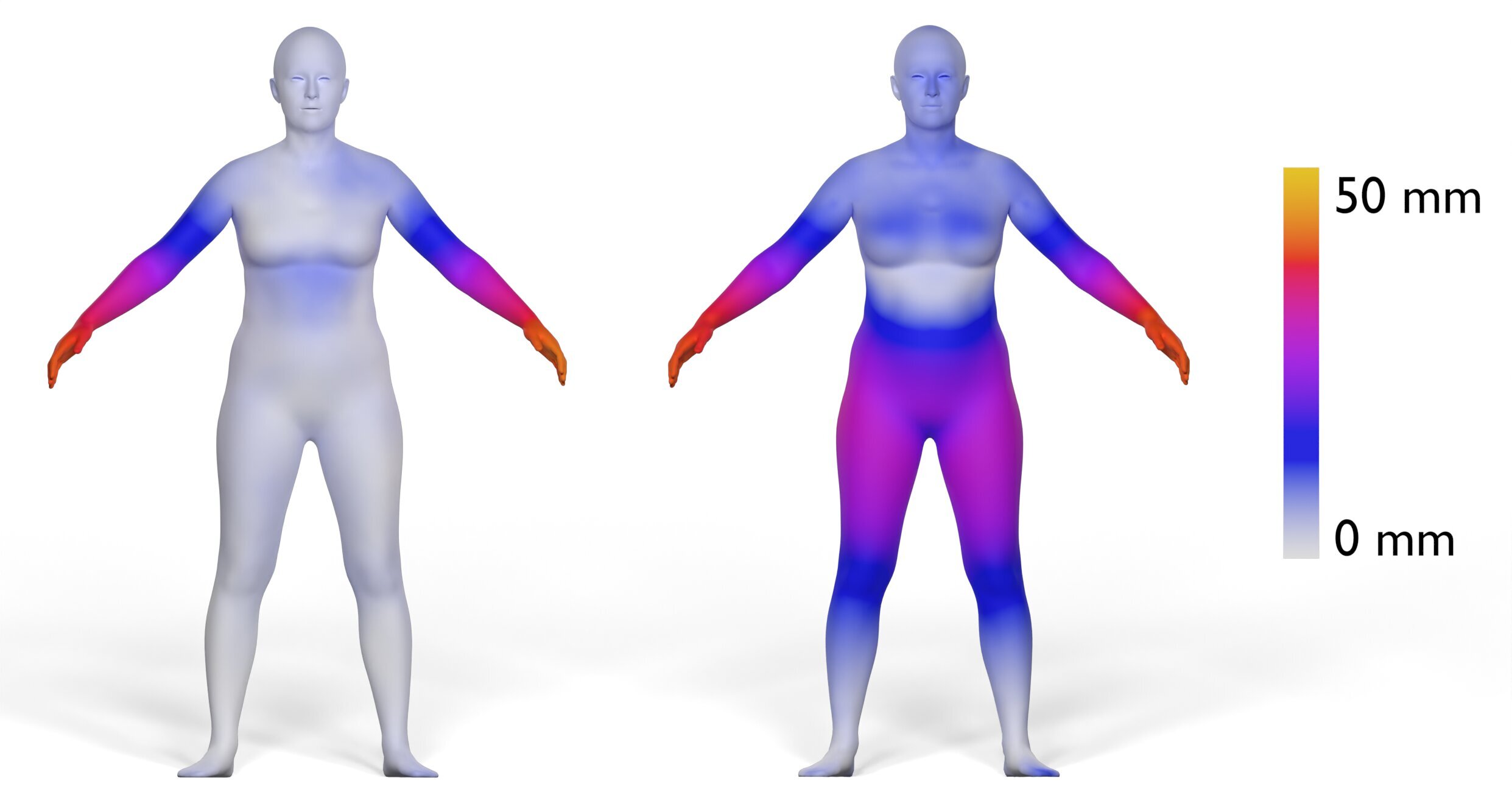}
    \caption{Comparison of our localized shape modification (left) with a global PCA approach \cite{piryankova2014} (right). Both models were used to shorten arm length by $\SI{38}{\mm}$. The vertex distance from the original to the modified mesh is color coded. Our model enables more localized shape changes, while the global PCA approach considerably changes the leg region when modifying arm length.}
    \label{fig:volumetric_vs_surface_pca}
\end{figure}

To show the benefits of our local and non-linear autoencoder and mesh convolution design (\Sec{learning}), we compare our method to the common approach of modelling anthropometric shape manipulation by correlating measurements with a global and linear PCA subspace learned from surface scans \cite{allen06learning, piryankova2014}. 
These methods allow global shape manipulation, but provide only limited \emph{local} control. 
For an example of this effect, we compare the results of shortening arm length with our model and the model proposed by Piryankova et al. \cite{piryankova2014}. 
\Fig{volumetric_vs_surface_pca} shows that our model provides local control of arm length, whereas the surface based approach results in notable changes in the leg region.
To interactively explore the shortcomings of the surface based approach, we invite the reader to experiment with the demo application at \url{https://bodyvisualizer.com}, and try to change the inseam parameter, while keeping the other measurements fixed.
This changes the model's arm length in addition to the desired effect of changing the leg length.


\subsection{Modifying Virtual Humans}
\label{sec:modifying_virtual_humans}

As demonstrated in \Fig{teaser}, we can also fit our model to surface scans of clothed humans, allowing us to modify virtual humans with our {\scshape TailorMe} model. 
To this end, given a registered surface scan conforming to our skin layer topology, we let our model infer skin and skeleton shape by optimizing the mean absolute error with additional weight decay in order to prevent overfitting.

To determine a fit for the skeleton and soft tissue distribution of a person, gradient descent is performed on the latent parameters $\vec{z}$ using the Adam optimizer \cite{kingma2015Adam} implemented in PyTorch and running on the GPU. 
We apply a weight decay of $7.5 \cdot 10^{-5}$ to prevent fitting values that are too far outside of the learned embeddings. 
Fitting the skeleton and soft tissue parameters takes $<\SI{100}{\ms}$ on a desktop PC equipped with an NVidia RTX 3090 GPU and an Intel Core i9 10850K CPU. 
The post-processing steps (\Sec{postprocessing}) add another \SI{700}{\ms} to the total inference time. 
This is an order of magnitude faster than the Inside Humans approach \cite{komaritzan2021inside}, whose authors report a total time of approximately $\SI{20}{\s}$ on similar hardware.
We measure an inference time of approximately $\SI{2}{\min}$ for the publicly available implementation of OSSO \cite{Keller2022OSSO}.

In order to modify the 3D scan of a person, we apply the changes made to the latent code as a delta shape manipulation to the scan of the person. 
By $\mat{g}_\theta(\tilde{\vec{z}})$ we denote the inference of our decoder for the fitted latent parameters $\vec{\tilde z}$ to a scan of a person $\mat{X}$. 
For modified latent parameters $\vec{z}$ we apply the difference of decoding $\tilde{\vec{z}}$ and $\vec{z}$  to the scan, resulting in the modified person $\mat{X}'$:
\begin{equation}
    \mat{X}' = \mat{X} + \left(\mat{g}_\theta(\vec{z}) - \mat{g}_\theta(\tilde{\vec{z}}) \right).
\end{equation}
To prevent unnatural deformation of the head region, we stitch the original head of the scanned person back onto the resulting mesh using differential coordinates similar to Döllinger et al.~\cite{dollinger2022resizeme}.

\section{Limitations}

Due to the limited availability of such data, our model is not trained on real anatomical data. 
We do note however, that the data needed for learning our model -- different soft tissue distributions on the same skeleton -- does not exist as ground truth data.
Recent methods have argued that relying on synthetic data alone could also be seen as an advantage and that the trained models can still outperform state of the art methods which are trained on real captured data \cite{Wood2021_FakeIt}.
However, evaluating our model on real anatomical data would still be desirable.

Our training data lacks information about the bone structure underlying the head, hands, and feet of our subjects. Therefore, our model cannot properly reproduce these areas. Due to this fact, we observe asymmetric structures, especially occurring in the facial region, which are amplified when modifying the latent code. 

Although our resulting meshes are free of self-intersections, this property only holds in the A-Pose of our template mesh. We have not done experiments on animating the resulting skeleton and skin.

\section{Conclusion}

We presented {\scshape TailorMe}, a novel approach for learning a volumetric anatomically constrained human shape model. 
We computed the Cartesian product of skeleton shapes and soft tissue distributions from the CAESAR database using volumetric deformation transfer. To decouple the two shape dimensions, we utilized a Barlow Twins inspired learning approach to train our autoencoder from pairs of skeleton and soft tissue distribution.
The resulting model can be used for shape sampling, e.g., generating various soft tissue distributions on the same skeleton. 
It provides localized shape manipulation due to the injected measurements in the latent space of our autoencoder. 
Compared to other methods, our model better decouples the skeleton and soft tissue shape dimensions, allows more localized shape manipulation, and provides significantly faster inference time.



In future work our model can be extended to include other anatomical details such as the muscles used to generate and modify humans. The skeleton, muscle, and soft tissue layers then can be separated by our model applying a triple Barlow twins loss, where pairs of eight are processed in a batch.
Incorporating a skeleton into the hands and feet will be beneficial rather than keeping them static. Fitting a skull inside the head, similar to the approach of Achenbach et al. \cite{achenbach2018}, enables a more accurate modification of the person's facial features. 

Being able to infer anatomical structures can be used to improve the animation of virtual humans.
Further investigations into developing an animation method for volumetric virtual humans that avoids self-intersections is an interesting direction for future work.




\bibliographystyle{unsrt}
\bibliography{references}

\end{document}